\documentclass[conference]{IEEEtran}
\IEEEoverridecommandlockouts
\usepackage{cite}
\usepackage{amsmath,amssymb,amsfonts}
\usepackage{algorithmic}
\usepackage{graphicx}
\usepackage{textcomp}
\usepackage{xcolor}
\usepackage{algorithm}
\usepackage{algorithmic}
\usepackage{booktabs}
\usepackage{multirow}
\usepackage{url}
\usepackage{balance}
\usepackage[colorlinks=true]{hyperref}
\def\BibTeX{{\rm B\kern-.05em{\sc i\kern-.025em b}\kern-.08em
    T\kern-.1667em\lower.7ex\hbox{E}\kern-.125emX}}
\begin{document}

\title{\Large \textbf{VLA-InfoEntropy: A Training-Free Vision-Attention Information Entropy \\ Approach for Vision-Language-Action Models Inference Acceleration and Success}}

\author{\IEEEauthorblockN{Chuhang Liu\textsuperscript{1,2}, Yayun He\textsuperscript{1}, Zuheng Kang\textsuperscript{1}, Xiaoyang Qu\textsuperscript{1}, Jianzong Wang\textsuperscript{1}\textsuperscript{$\dagger$}}
\IEEEauthorblockA{
$^1$Ping An Technology (Shenzhen) Co., Ltd., Shenzhen, China \\$^2$Tsinghua Shenzhen International Graduate School, Tsinghua University, Shenzhen, China 
}
\thanks{
 Supported by Shenzhen-Hong Kong Joint Funding Project (Category A) under grant No. SGDX20240115103359001. \textsuperscript{$\dagger$}Corresponding author is Jianzong Wang (jzwang@188.com)
 } 
}

\maketitle

\begin{abstract}
    Vision-Language-Action (VLA) models integrate visual perception, language understanding, and action decision-making for cross-modal semantic alignment, exhibiting broad application potential. However, the joint processing of high-dimensional visual features, complex linguistic inputs, and continuous action sequences incurs significant computational overhead and low inference efficiency, thereby hindering real-time deployment and reliability.
    To address this issue, we use image entropy to quantify the grayscale distribution characteristics of each visual token and introduce attention entropy to capture the distribution of attention scores over task-related text. Visual entropy identifies texture-rich or structurally informative regions, while attention entropy pinpoints semantically relevant tokens. Combined with timestep information, these metrics enable a dynamic transition strategy that shifts the model's focus from global visual features to attention-guided local informative regions.
    Thus, the resulting VLA-InfoEntropy method integrates spatial, semantic, and temporal cues to reduce redundancy while preserving critical content. Extensive experiments show that our method reduces inference parameters, accelerates inference speed, and outperforms existing approaches.
\end{abstract}


\section{Introduction}
\label{sec:intro}

Vision–Language–Action (VLA) models integrate visual and linguistic information to enable end-to-end decision-making and have shown strong potential in applications such as autonomous driving, robotics, and intelligent assistants\cite{sun2024comprehensive, duan2022survey, wang2025large, zheng2025survey}. 
However, the high computational cost of VLA inference remains a major barrier to large-scale and real-time deployment, where inference latency directly affects system reliability and responsiveness\cite{yao2025advancing, firoozi2025foundation, ren2024embodied}.

\begin{figure}[t!] 
    \centering 
    \includegraphics[width=\linewidth]{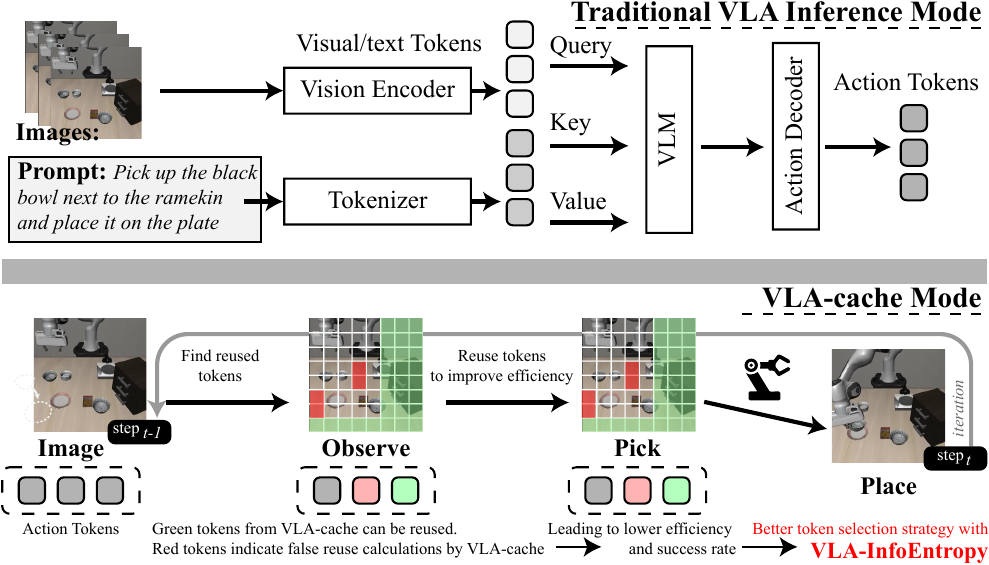} 
    \caption{\textbf{The motivation of VLA-InfoEntropy}. In traditional VLA inference mode, all tokens are computed indiscriminately. Although this ensures full coverage, many tokens are clearly redundant. The VLA-cache mode\cite{xu2025vlacache} addresses this issue, but its token selection strategy requires further optimization. To this end, we propose the VLA-InfoEntropy method, which identifies the most important tokens in an image, excludes them from the static token set, and then reuses the key-value (KV) cache for efficient computation.} 
    \label{fig:res1}
\end{figure}

Recent studies have attempted to improve efficiency by reducing redundancy in visual, linguistic, and action representations, as VLA inference frequently processes highly repetitive information. LightVLA\cite{jiang2025better} dynamically prunes tokens via cross-attention and Gumbel-softmax, FLASH-VLA~\cite{tan2025think} eliminates low-value visual tokens using action similarity, VLA-Cache~\cite{xu2025vlacache} reuses static visual tokens through a layer-adaptive KV cache, and SP-VLA~\cite{li2026spvla} applies action-aware scheduling with spatiotemporal token pruning.

However, these methods typically address visual redundancy, semantic relevance, and temporal coherence in isolation, limiting their effectiveness in sequential decision-making. Visual pruning may remove instruction-critical cues, while attention-based pruning can be misled by noise, focusing on irrelevant regions. Additionally, without timestep information, tokens remain fixed throughout the rollout, preventing the model from adapting to task-relevant details~\cite{jung2025consistency}.

To address these issues, we propose \textbf{VLA-InfoEntropy}, a training-free inference acceleration framework that jointly leverages visual, semantic, and temporal information.
The method introduces two entropy-based metrics—visual entropy and attention entropy—and incorporates timestep-aware selection to dynamically characterize information distribution and task relevance. This unified strategy enables adaptive allocation of computational resources during inference, preserving decision-critical cues while substantially reducing redundant computation, thereby achieving efficient and flexible multimodal reasoning. Overall, our contributions are as follows:

\begin{figure*}[t!] 
  \centering 
  \includegraphics[width=\textwidth]{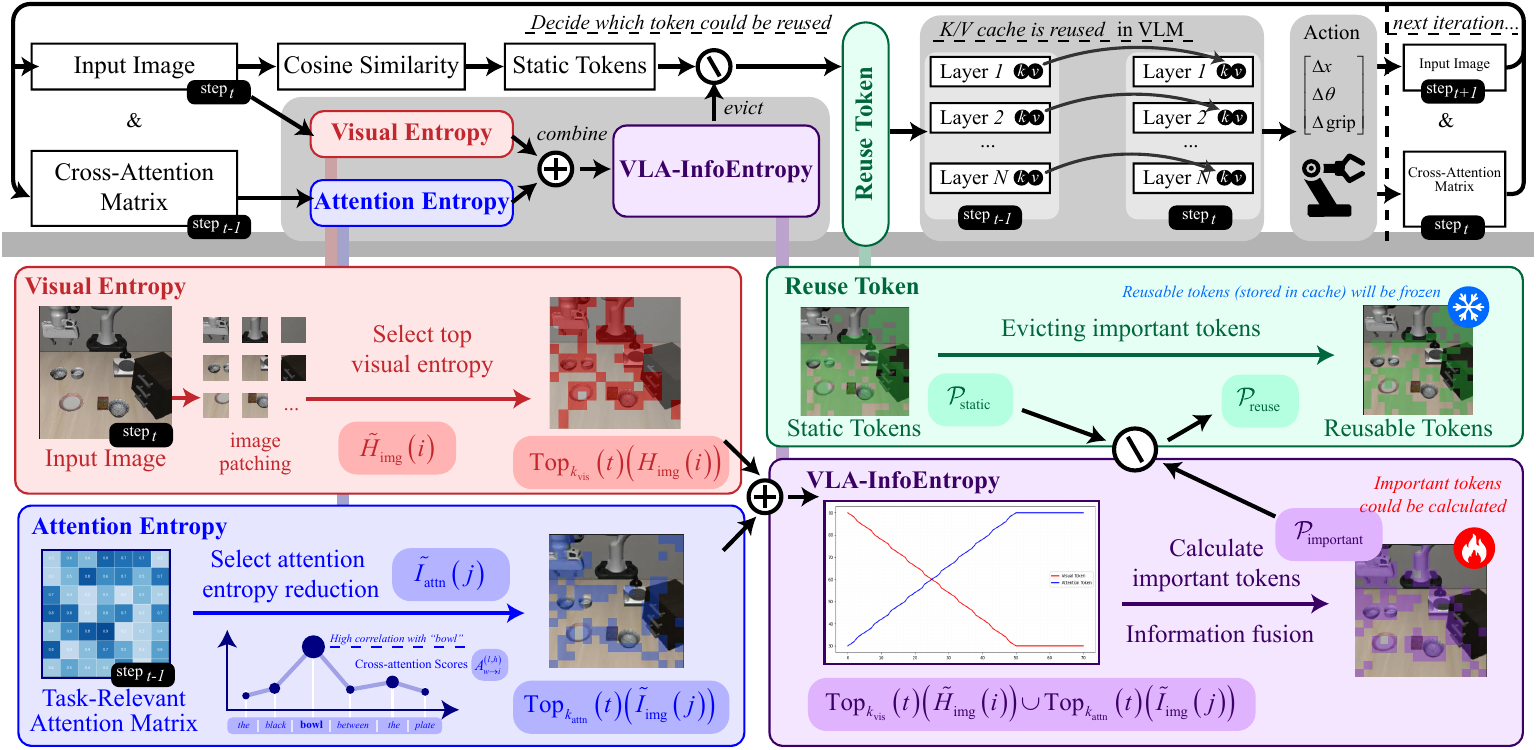} 
  \caption{\textbf{The overview of the proposed method.}
  \textbf{Visual Entropy}: Compute visual entropy for each image token to quantify its intrinsic information content. Tokens from texture-rich and edge-dense regions are selected as visually salient candidates.
  \textbf{Attention Entropy}: Extract text–vision attention scores from the cross-attention matrix, compute attention entropy, and identify tokens whose attention is strongly concentrated on the instruction, marking them as task-relevant.
  \textbf{VLA-InfoEntropy}: Using the timestep $t$, dynamically adjust the proportions of tokens selected by visual and attention entropy, gradually shifting the focus from global visual context at early steps to task-driven local evidence at later steps.
  \textbf{Reuse token}: important tokens are removed from the static token set, and the remaining low-information tokens are treated as reusable KV-cache tokens, enabling efficient inference while suppressing redundant computation.}
  \label{fig:res2} 
\end{figure*}

\begin{itemize}
\item We introduce two complementary, training-free entropy metrics: \emph{visual entropy}, which quantifies the intrinsic information content of visual tokens, and \emph{attention entropy}, which captures their semantic relevance to task-related language. These measures are integrated into a unified token selection framework.
\item We introduce \textbf{VLA-InfoEntropy}, which further incorporates timestep-aware modeling to progressively transition the model’s focus from global visual context to task-specific semantic cues, enabling efficient spatial–semantic–temporal token selection that improves inference efficiency while maintaining strong task performance. In conclusion, our key contribution is a novel multimodal fusion strategy that provides a principled way to quantify multimodal information importance.
\end{itemize}

\section{Related Work}

\subsection{Vision-Language-Action Models}

With the rapid advancement of multimodal large models, Vision-Language Models (VLMs) have been extended to generate actions. RT-1\cite{brohan2023rt1} discretizes actions using a Transformer-based framework, while RT-2\cite{zitkovich2023rt} introduces the concept of VLA, leveraging VLMs' prior knowledge for decision-making. Researchers have also explored diffusion-based approaches\cite{chi2025diffusion, wen2025diffusionvla, black2025pi0, black2025pi05, bjorck2025gr00t} to facilitate smoother action transitions. Additionally, combining world models\cite{ding2025understanding, zhang2025dreamvla} with VLA frameworks enables the prediction of future frames, improving the system's ability to understand and interact with dynamic environments. Despite these advances, VLA systems still require significant computational resources, limiting their scalability and real-time deployment.

\subsection{Acceleration for Vision-Language-Action Models}

To address the computational bottleneck of VLA models, numerous training-based acceleration methods have been proposed: OpenVLA-OFT\cite{kim2025finetuningvla} and PD-VLA\cite{song2025pd}, which focus on parallel decoding and action chunking, FAST\cite{pertsch2025fast} for DCT-based tokenization, and CEED-VLA\cite{song2025ceed} for consistency distillation, all of which contribute to improving acceleration. However, these methods require expensive computational resources. 
In contrast, training-free methods\cite{zhang2025sparsevlm, chen2024image, xu2025vlacache, li2026spvla, yang2025efficientvla} accelerate inference without modifying model weights. These methods include techniques such as pruning redundant visual tokens, optimizing attention patterns, and reusing intermediate features.
However, most existing approaches treat visual and attention processing separately and ignore temporal dependencies\cite{yue2024deer}.
Our work addresses this gap to achieve more efficient VLA inference.

\section{Methodology}

In VLA inference, efficiency is hindered by repeatedly processing visual tokens that contain little useful information. VLA-Cache~\cite{xu2025vlacache} mitigates this redundancy by leveraging attention scores across visual tokens and text ($\S$~\ref{subsec:cache_vla}), but attention alone is an incomplete and inefficient criterion. A more effective measure should capture both visual complexity and task relevance.
Entropy offers a training-free, theoretically grounded metric for token informativeness~\cite{shannon1948mathematical, kapur1985new}. Visually complex regions exhibit higher entropy, while uniform backgrounds tend to have lower entropy and are often redundant (vision entropy, $\S$~\ref{subsec:two_entropy}). Similarly, cross-modal attention distributions reflect semantic relevance: task-relevant visual tokens yield concentrated, low-entropy attention, whereas irrelevant tokens produce diffuse, high-entropy attention (attention entropy, $\S$~\ref{subsec:two_entropy}). Combining these two entropy measures enables a unified criterion that captures both spatial and semantic importance, supporting a time-dependent inference strategy that shifts from global context to task-specific details ($\S$~\ref{subsec:vlaie}).

\subsection{Cache for Vision-Language-Action Models}
\label{subsec:cache_vla}

The key idea of VLA-Cache~\cite{xu2025vlacache} is to partition the images into patches and compute the cosine similarity between corresponding patches across frames. This allows us to identify a set of static tokens $\mathcal{P}_{\text{static}}$ , which exhibit low similarity over time. Crucially, we exclude certain important tokens from this set, which defines the set of tokens that cannot be ignored. The remaining reusable tokens are captured in the set $\mathcal{P}_{\text{reuse}} = \mathcal{P}_{\text{static}} \setminus \mathcal{P}_{\text{important}}$, which are eligible for KV-cache reuse.
For each visual token $i$:
\begin{equation}
\label{eq:1}
    \begin{cases}
        K_{t}(i), V_{t}(i) = K_{t-1}(i), V_{t-1}(i), & \text{if} \quad i\in \mathcal{P}_{\text{reuse}} \\
        K_{t}(i),V_{t}(i) =W_{K}X_{t}(i),W_{V}X_{t}(i), & \text{Otherwise}.
    \end{cases}
\end{equation}

Thus, excluding important tokens from the static set is critical for maintaining a balance between inference acceleration and model performance.

\subsection{Visual Entropy and Attention Entropy}
\label{subsec:two_entropy}

Image entropy measures the randomness of pixel intensities, with more dispersed grayscale values indicating higher entropy and richer texture. We estimate visual information based on grayscale distribution, assuming that tokens with complex textures or clear edges carry more important cues. Thus, image entropy $H_{\text{img}}$ helps identify key tokens.
\begin{equation}
\label{eq:2}
    H_{\text{img}}(i) = -\sum_{g=0}^{G-1} p_i(g) \log_2 p_i(g),
\end{equation}
where $i$ indexes the $i$-th visual token (out of 256), and $G$ (typically 256) is the number of grayscale levels. For each token $i$, we compute its grayscale histogram ${p_i(g)}_{g=0}^{G-1}$, where $p_i(g)$ is the probability of gray level $g$ in the corresponding region.

To facilitate unification with the subsequent attention entropy, we apply maximum-entropy normalization as $\widetilde{H}_{\text{img}}$:
\begin{equation}
\label{eq:3}
    \widetilde{H}_{\text{img}}(i) = \frac{H_{\text{img}}(i)}{\log_2 G},
\end{equation}

Visual tokens unrelated to task prompts lead to dispersed distributions of attention scores and higher attention entropy, whereas related tokens result in concentrated distributions and lower attention entropy. Based on this, we define the attention distribution as follows:
\begin{equation}
\label{eq:4}
    q_(w,i)= \frac{
    \exp \left(\frac{1}{LH}\sum_{l=1}^{L}\sum_{h=1}^{H}A^{(l,h)}_{w\to i} \right)}{
    \sum_{w'\in\mathcal{W}} \exp\left(\frac{1}{LH}\sum_{l=1}^{L}\sum_{h=1}^{H}A^{(l,h)}_{w'\to i} \right)
    },
    w \in \mathcal W,
\end{equation}
where \(A^{(l,h)}\) denotes the cross-attention matrix (after softmax) at layer \(l\) and head \(h\), which contains attention scores between all text and visual tokens; \(w\) indexes an individual text token, and \(\mathcal W\) denotes the set of all text tokens in the input sequence. For each visual token \(i\), we extract the text-to-vision entries \(A^{(l,h)}_{w \to i}\) for all \(w \in \mathcal W\) and average these entries over all layers and heads. Applying a softmax operation along the text-token dimension yields \(q_{(w,i)}\), the normalized distribution of visual token \(i\) over the text token set \(\mathcal W\).

\begin{table*}[t] 
\centering 
\caption{Experimental results of different models on the LIBERO benchmark. Boldface indicates the best results.} 
\label{TAB1}
\resizebox{\textwidth}{!}{   
\begin{tabular}{lccccc ccc} 
\toprule 
\multirow{2}{*}{Method} & 
\multicolumn{5}{c}{Success Rate~(\%,$\uparrow$)} & 
\multirow{2}{*}{Latency~($\downarrow$)} &
\multirow{2}{*}{FLOPs~($\downarrow$)} &
\multirow{2}{*}{Speed up~($\uparrow$)} \\ 
\cmidrule(lr){2-6} 
& LIBERO-Spatial & LIBERO-Object & LIBERO-Goal & LIBERO-Long & Average &  &  \\
\midrule 
OpenVLA\cite{kim2025openvla} (baseline) & 84.4$\pm$0.9 & 86.6$\pm$0.8 & 75.6$\pm$1.0 & 53.2$\pm$1.4 & 75.0$\pm$0.6 & 51.91 & 1.864 & 1.00 \\
SparseVLM\cite{zhang2025sparsevlm} & 79.8$\pm$0.6 & 67.0$\pm$1.2 & 72.6$\pm$1.4 & 39.4$\pm$0.8 & 64.7$\pm$1.0 & 83.39 & 1.407 & 1.32 \\
FastV\cite{chen2024image}  & 83.4$\pm$1.2 & 84.0$\pm$0.6 & 74.2$\pm$1.4 & 51.6$\pm$0.8 & 73.3$\pm$1.1 & 53.28 & 1.864 & 1.00  \\
VLA-Cache\cite{xu2025vlacache} & 83.8$\pm$0.9 & 85.8$\pm$0.7 & 76.4$\pm$0.6 & 52.8$\pm$1.3 & 74.7$\pm$0.8 & 34.38 & 1.355 & 1.38 \\
SP-VLA\cite{li2026spvla} & 84.4$\pm$1.0 & 85.6$\pm$1.6 & 75.4$\pm$1.8 & 54.2$\pm$1.4 & 74.9$\pm$1.3 & - & 3.100  & 1.35 \\
Spec-VLA\cite{wang2025spec} & 85.8$\pm$0.8 & 85.0$\pm$2.0 & 74.4$\pm$1.4 & \textbf{55.0$\pm$1.2} & 75.0$\pm$1.4 & - & - & 1.32 \\
\midrule
\textbf{Ours} & \textbf{86.4$\pm$1.0} & \textbf{87.6$\pm$1.4} & \textbf{79.4$\pm$0.8} & 52.2$\pm$1.0 & \textbf{76.4$\pm$1.0} & \textbf{31.25} & \textbf{1.214} & \textbf{1.53} \\
\bottomrule 
\end{tabular} 
}
\end{table*} 

Based on this, we can further define the attention entropy for the $i$-th visual token, which quantifies the uncertainty of the model’s attention distribution $H_{\text{attn}}$, as follows:
\begin{equation}
\label{eq:5}
    H_{\text{attn}}(i) = -\sum_{w\in\mathcal W} q(w,i)\log_2 q(w,i).
\end{equation}

Accordingly, we can define the ``amount of information'' based on the entropy reduction (i.e., mutual information) relative to a uniform prior, and we normalize this value as:
\begin{equation}
\label{eq:6}
    \widetilde{I}_{\text{attn}}(i)=1-\frac{H_{\text{attn}}(i)}{\log_2 |\mathcal W| }.
\end{equation}

\subsection{VLA-InfoEntropy}
\label{subsec:vlaie}
Inspired by the human visual perception process—which first captures global visual information and then shifts attention to local details—we design a novel dynamic selection strategy:
\begin{equation}
\label{eq:7}
\begin{cases}
    k_{\text{vis}}(t) = \lfloor k_2 - ( k_2 - k_1 ) \cdot \alpha \rfloor  \\  
    k_{\text{attn}}(t) = \lfloor k_1 + ( k_2 - k_1 ) \cdot \alpha \rfloor,
\end{cases}
\end{equation}
where $k_1$ , $k_2$ are two hyperparameters denoting the number of visual tokens and task-related tokens, respectively. The transition between these two allocation schemes occurs over $T$ steps, governed by the action execution timestep $t$. The temporal parameter is defined as \space $\alpha = \dfrac{t}{T} \in [0,1]$.

Building on this dynamic selection strategy, we proposed \textbf{VLA-InfoEntropy}, which is expressed as follows:
\begin{equation}
\label{eq:8}
    \mathcal{P}_{\text{important}}(t) = \text{Top}_{k_{\text{vis}}}(t)\bigl(\widetilde{H}_{\text{img}}(i)\bigr) 
    \cup 
    \text{Top}_{k_{\text{attn}}}(t)\bigl(\widetilde{I}_{\text{attn}}(j)\bigr),
\end{equation}
where $i$ and $j$ denote visual tokens.

By selecting tokens with the highest visual and attention entropy scores and applying a timestep-dependent mechanism, our method adaptively identifies the most important tokens during inference. This method achieves a unified integration of visual, semantic, and temporal cues, thereby enabling more effective multimodal reasoning. The upper part of Figure~\ref{fig:res2} shows the overall pipeline, while the lower part explains the function and implementation of each module. Full details of the algorithm, along with computational complexity and resource consumption, are presented in the appendix.

\begin{figure*}[t!]
  \centering
  \includegraphics[width=1.0\textwidth]{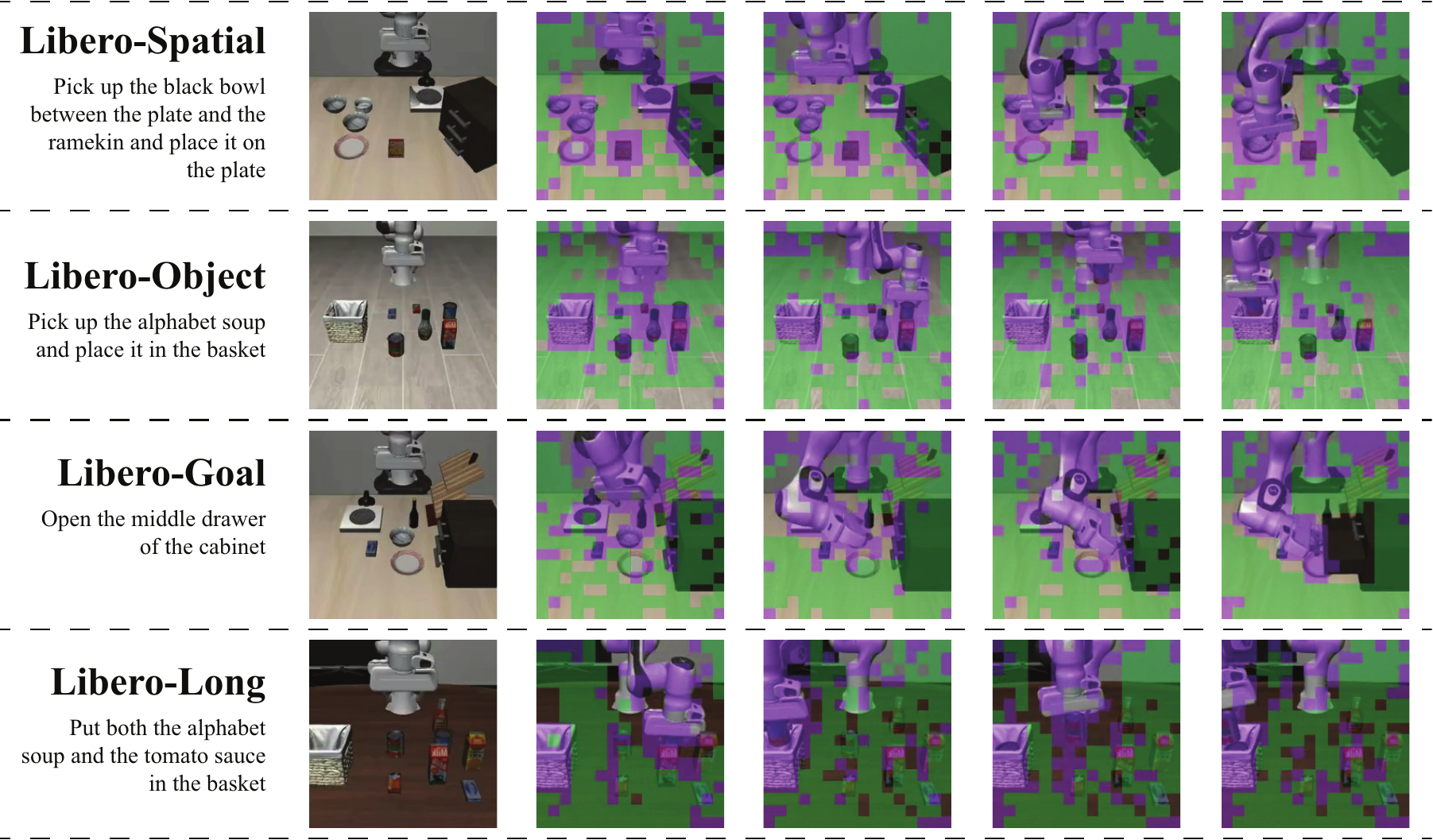}
  \caption{\textbf{The visualization results of the LIBERO}. For representative tasks from LIBERO-Spatial, LIBERO-Object, LIBERO-Goal, and LIBERO-Long, the figure illustrates how VLA-InfoEntropy highlights task-relevant regions while suppressing redundant background areas. Important tokens (purple) tend to cluster around manipulable objects and interaction points, whereas reusable tokens (green) are distributed over uninformative regions. These results demonstrate that the proposed entropy-based token selection effectively concentrates computation on task-relevant visual evidence, thereby reducing overall computational cost while accelerating inference.} 
  \label{fig:res3} 
\end{figure*}

\begin{table*} 
\centering 
\caption{Ablation Study. The importance of visual entropy, attention entropy, and timestep are evaluated separately.} 
\label{TAB2}
\resizebox{\textwidth}{!}{   
\begin{tabular}{lccccc ccc} 
\toprule 
\multirow{2}{*}{Method} & 
\multicolumn{5}{c}{Success Rate~(\%,$\uparrow$)} &
\multirow{2}{*}{Latency~($\downarrow$)} &
\multirow{2}{*}{FLOPs~($\downarrow$)} &
\multirow{2}{*}{Speed up~($\uparrow$)} \\
\cmidrule(lr){2-6} 
& LIBERO-Spatial & LIBERO-Object & LIBERO-Goal & LIBERO-Long & Average &  &  \\
\midrule 
OpenVLA(baseline) & 84.4 & 86.6 & 75.6 & \textbf{53.2} & 75.0 & 51.91 & 1.864 & 1.00 \\
Visual-Entropy  & 82.0 & 82.6 & 73.8 & 42.6 & 69.8 & 33.82 & 1.378 & 1.42 \\
Attention-Entropy & 83.4 & 87.6 & 76.4 & 50.2 & 73.9 & 33.24 & 1.311 & 1.39 \\
Visual+Attention-Entropy & 85.4 & 86.6 & 79.2 & 51.8 & 75.8 & 32.26 & 1.244 & 1.49  \\
\midrule
\textbf{Ours} & \textbf{86.4} & \textbf{87.6} & \textbf{79.4} & 52.2 & \textbf{76.4} & \textbf{31.25} & \textbf{1.214} & \textbf{1.53} \\
\bottomrule 
\end{tabular} 
}
\end{table*} 

\section{Experiments}

\subsection{Datasets}

To validate our method, we adopt LIBERO~\cite{liu2023libero}—a widely used benchmark for VLA model evaluation—comprising four task suites: LIBERO-Object, LIBERO-Spatial, LIBERO-Goal, and LIBERO-Long. Covering variations in objects, spatial relations, goals, and long-horizon scenarios, it assesses VLA models’ generalization capabilities in semantic grounding, spatial reasoning, goal understanding, and multi-step decision-making, making it a comprehensive benchmark for evaluating VLA systems’ competence and robustness.

\subsection{Experimental Settings}

The cross-attention matrix $A^{(l,h)}$ is taken from the computation of the last frame, a choice that introduces no statistically significant performance degradation\cite{xu2025vlacache}. 
The dynamic schedule adopts $T=100$, enabling a gradual transition from global visual guidance to local attention focus across the $100$-step rollout. We set $k_1=40$ and $k_2=60$ to balance spatial and semantic information during token selection. All experiments are conducted on a single NVIDIA RTX 4090 GPU; results may vary under different hardware or runtime conditions. All experiments are repeated multiple times.

\subsection{Main Results}

Table~\ref{TAB1} summarizes the main experimental results, comparing VLA-InfoEntropy with the OpenVLA baseline~\cite{kim2025openvla} and several state-of-the-art methods. Overall, the results show that VLA-InfoEntropy significantly improves inference efficiency while maintaining strong task performance. On LIBERO-Spatial, it achieves a success rate of $86.4\%$, outperforming all competing approaches, with an average success rate of $76.4\%$ across tasks. In terms of efficiency, VLA-InfoEntropy delivers a $1.53\times$ speedup in FLOPs, accompanied by a $34.9\%$ reduction in FLOPs and a $39.8\%$ decrease in CUDA latency under the evaluated hardware settings. LIBERO-Long is challenging due to long horizons and error accumulation, yet our method maintains competitive performance while reducing computational overhead, which is crucial for real-time embodied deployment.

These findings indicate that conventional VLA inference pipelines contain substantial informational redundancy, which can dilute the model’s attention, introduce spurious correlations, and degrade both efficiency and success rates. By contrast, the entropy-guided selection in VLA-InfoEntropy preserves high-information, task-relevant multimodal cues while suppressing redundant or misleading features, resulting in more compact evidence usage and reduced error propagation. The consistent gains across diverse benchmarks demonstrate that redundancy reduction via entropy-based filtering can substantially lower computational cost and latency without sacrificing—and in some cases improving—task performance. Qualitative visualizations illustrating how VLA-InfoEntropy differentiates task-critical from redundant visual regions are shown in Fig.~\ref{fig:res3}.  Additional visualizations are provided in the appendix.

\subsection{Ablation Study}
With the number of selected important tokens fixed at $100$, we evaluate four settings: image entropy only, attention entropy only, a static combination, and the proposed dynamic VLA-InfoEntropy. This study isolates the effects of visual cues, semantic relevance, and temporal dynamics. As shown in Table~\ref{TAB2}, visual-only selection misses task semantics, while attention-only selection ignores important visual information. A static combination improves performance but fails to model temporal variation. VLA-InfoEntropy, which incorporates timestep-conditioned selection, achieves the superior results.

\begin{figure}[t!] 
  \centering 
  \includegraphics[width=\linewidth]{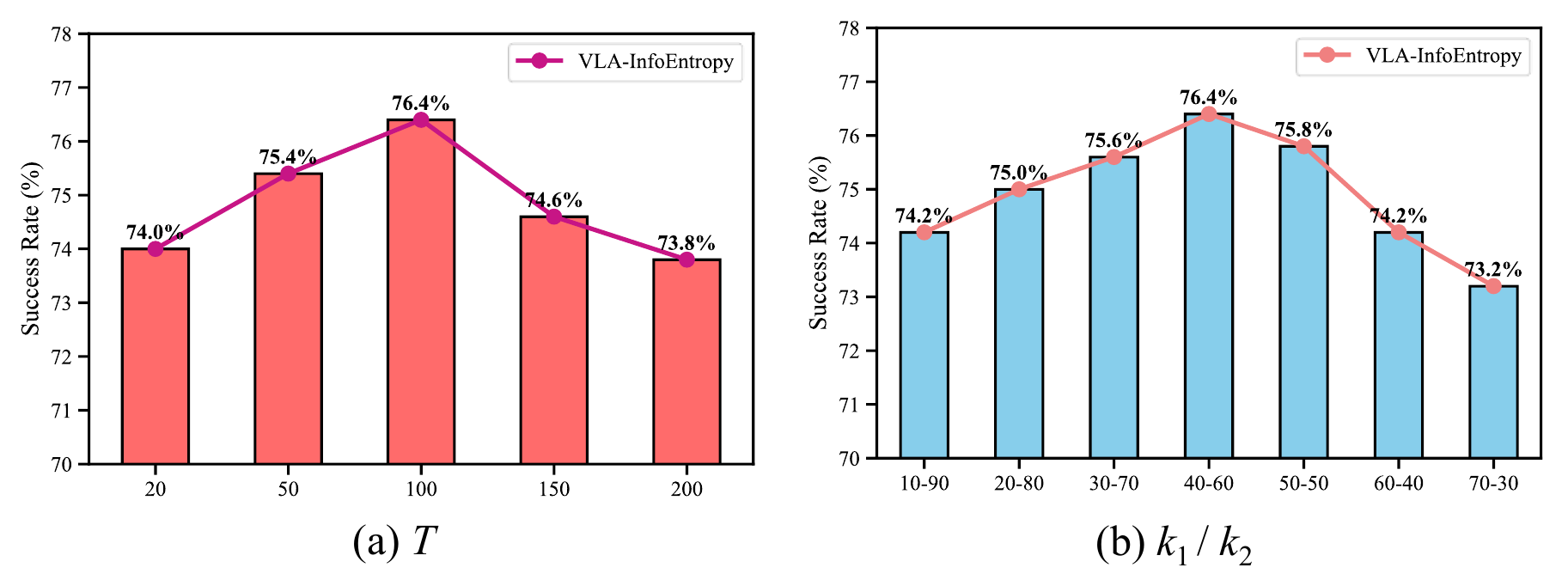} 
  \caption{\textbf{Hyperparameter Sensitivity.} \textbf{(a)} The success rate initially rises and then declines as the hyperparameter $T$ increases. \textbf{(b)} The success rate also increases at first and then decreases as the ratio of the hyperparameters $k_1/k_2$ increases.} 
  \label{fig:res4} 
\end{figure}

\subsection{Sensitivity Analysis}

We analyze sensitivity to key hyperparameters using average success rates across four benchmarks. As shown in Fig.~\ref{fig:res4}, performance peaks at $T=100$. We then vary the ratio $k_1/k_2$ to balance visual and semantic tokens, with spatial tasks favoring visual cues and goal tasks favoring semantics; the optimal ratio is $40$–$60$. Finally, fixing $k_1/k_2=2{:}3$, increasing the total number of tokens $k_1+k_2$ substantially increases latency (Table~\ref{Tab3}), while success rates saturate beyond $100$ tokens. These results emphasize the need to balance efficiency and visual–semantic allocation.

\begin{table}[t!]
\centering
\caption{Comparison of different total number of important tokens}  
\label{Tab3}
\resizebox{\linewidth}{!}{
\begin{tabular}{c|c|c|c} 
\toprule
\multicolumn{1}{c|}{Tokens}  & \multicolumn{1}{c|}{Success Rate (\%,$\uparrow$)} & \multicolumn{1}{c|}{Latency ($\downarrow$)} & \multicolumn{1}{c}{FLOPs ($\downarrow$)} \\ 
\midrule  
60 & 71.6 & \textbf{30.89} & \textbf{1.108}  \\
80 & 75.2 & 31.00 & 1.157  \\
100 & 76.4 & 31.25 & 1.205 \\
120 & 75.4 & 34.49 & 1.251 \\
140 & \textbf{76.8} & 36.64 & 1.297 \\
\bottomrule
\end{tabular}
}
\end{table}

\section{Conclusion}

This paper presents VLA-InfoEntropy, a training-free dynamic inference framework that evaluates multimodal information using image entropy and attention entropy. By enabling a smooth transition from global visual cues to locally focused attention, the method reduces redundant computation while preserving essential representations. Combined with the KV reuse mechanism of VLA-Cache, this unified information-theoretic approach allows efficient retention of static or non-critical tokens.
Experiments on LIBERO show that VLA-InfoEntropy reduces inference latency and FLOPs while maintaining or improving task performance, achieving a 76.4\% average success rate, 34.9\% fewer FLOPs, and 39.8\% lower CUDA latency. 
These findings expose substantial redundancy in standard VLA pipelines and their negative impact on inference efficiency. Our method offers a training-free solution for real-time VLA acceleration by integrating visual and attention-based signals to suppress redundant evidence, underscoring the importance of information fusion and redundancy mitigation in embodied intelligence.

\bibliographystyle{IEEEbib}
\balance
\bibliography{refs}

\end{document}